\documentclass{article}

\usepackage[final]{nips_2017}

\usepackage[utf8]{inputenc} % allow utf-8 input
\usepackage[T1]{fontenc}    % use 8-bit T1 fonts
\usepackage{hyperref}       % hyperlinks
\usepackage{url}            % simple URL typesetting
\usepackage{booktabs}       % professional-quality tables
\usepackage{amsfonts}       % blackboard math symbols
\usepackage{nicefrac}       % compact symbols for 1/2, etc.
\usepackage{microtype}      % microtypography
\usepackage{graphicx}
\usepackage{subcaption}

\title{Learning to Compose Skills}

\author{
  Himanshu Sahni \\
  Georgia Institute of Technology \\
  \texttt{himanshu@gatech.edu} \\
  \And
  Saurabh Kumar\thanks{Denotes equal contribution.} \\
  Georgia Institute of Technology \\
  \texttt{skumar311@gatech.edu} \\
  \AND
  Farhan Tejani\footnotemark[1] \\
  Georgia Institute of Technology \\
  \texttt{farhantejani@gatech.edu} \\
  \And
  Charles Isbell \\
  Georgia Institute of Technology \\
  \texttt{isbell@cc.gatech.edu} \\
}

\begin{document}

\maketitle

\begin{abstract}
We present a differentiable framework capable of learning a wide variety of compositions of simple policies that we call skills. By recursively composing skills with themselves, we can create hierarchies that display complex behavior. Skill networks are trained to generate skill-state embeddings that are provided as inputs to a trainable composition function, which in turn outputs a policy for the overall task. Our experiments on an environment consisting of multiple collect and evade tasks show that this architecture is able to quickly build complex skills from simpler ones. Furthermore, the learned composition function displays some transfer to unseen combinations of skills, allowing for zero-shot generalizations.

\end{abstract}

\section{Introduction}
A key property of intelligent agents is the ability to learn simple skills throughout their lifetimes and compose them together to solve complicated tasks. Yet, traditional reinforcement learning (RL) agents lack this ability, making it hard to learn in environments with long term dependencies. Recent advances in using deep neural networks as function approximators allow for learning in high dimensional state spaces \cite{MnihAtari}, but do not address this fundamental problem. Hierarchical RL \cite{DietterichHierarchical, vezhnevets2017feudal} offers a solution by decomposing a single complicated task into a hierarchy of simpler subtasks, often times using intrinsic rewards to motivate underlying learners. A related strategy is to use \textit{options} \cite{SuttonOptions}, a set of policies with fixed, possibly stochastic, initiation and termination criteria, that are made available to the agent along with base environment actions. Both approaches focus on decomposing a difficult problem into a \textit{sequence} of simpler subgoals. The motivation behind this work is that solving problems in the real world rarely calls for optimal sequential decompositions of arbitrary tasks; instead, a set of basic skills can be composed in \textit{multiple} interesting ways to exhibit complex behavior.

A major distinction between our work and recent attempts to learn an optimal sequence of subgoals \cite{andreas2016modular, oh2017zero, frans2017meta} is that our framework can learn a much richer set of compositions of skills. For example, in the game of Pacman, an agent must learn to collect food pellets while also avoiding enemy ghosts. In the usual view of hierarchical RL, a subgoal or option, such as "navigate to food pellet A" or "evade enemy ghost", would be activated one at a time and the agent must learn to alternate between them to complete the overall task. A better approach is to learn a policy that composes both subgoals, i.e. identifies food pellets that also keep Pacman far away from ghosts and prioritizes their collection. In this work, we consider a subset of compositions defined by Linear Temporal Logic (LTL) \cite{pnueli1992temporal, baier2008principles}. A wide variety of common RL tasks can be specified using the temporal modal operators defined in LTL: next ($\mathcal{O}$), always ($\square$), eventually ($\lozenge$), and until ($\mathcal{U}$), along with the basic logic connectives: negation($\neg$), disjunction ($\vee$), conjunction ($\wedge$) and implication ($\rightarrow$) \cite{littman2017environment}. The Pacman task above can be translated into LTL as $\neg G \; \mathcal{U} \; (\lozenge F_{1} \, \wedge \, \lozenge F_{2} \, \wedge \, \ldots \, \lozenge F_{n})$, where $G$ is the proposition that the Pacman occupies the same location as any of the ghosts and $F_{1}$ through $F_{n}$ are the corresponding propositions for all the food pellets. Thus, the LTL sentence can be interpreted as ``do not get eaten by a ghost until all the food pellets have been collected''.

Our main contribution in this work is the expression of these compositions as differentiable functions. Representations of the individual skill policies are fed to this function as inputs and a representation for the composed task policy is produced. Skill policies are learned only once, and a wide variety of compositions can be created after the fact. We show that learning to compose skills is more efficient than learning to sequence those skills as is typically done in hierarchical RL. Moreover, we show how recursive compositions can be used to create rich hierarchies for more complicated behavior. 

A challenge with trainable compositions is that skill policies must be represented in a differentiable manner so that they can be utilized inside the composition function. In most modern RL domains, policies are represented as deep neural networks, with the outputs normalized to form a probability distribution over actions. The action distribution alone, however, may not encode enough information on the importance of a subgoal in the current state to arbitrate between competing subgoals. On the other hand, the entire policy network may contain multiple layers and thousands of weights. Trying to learn a composition function over that would be very challenging. Therefore, we use a special architecture for training skill policies which allows us to embed information on the skill and the state in a single layer of a network. We call these skill-state embeddings. Each embedding is then fed into a composition layer which learns to solve the overall task (see section \ref{sec:architecture}). The cost of acquiring skills is one time and low, and training the composition function is faster than learning the overall task from scratch. More importantly, the skills can be reused over and over again for different compositions. Finally, we show that the composition function itself shows some transfer to unseen tasks, allowing for zero-shot task generalization. 

\section{Background}
\label{background}
In RL, an agent interacts with a dynamic environment and learns to maximize the notion of a long term reward. The task is typically characterized as a Markov Decision Process (MDP) defined by the tuple $\{S, A, T, R, \gamma\}$ of states, actions, transition function, reward, and discount factor. A policy $\pi(S) \rightarrow P(A)$ maps each state to a probability distribution over actions that the agent should take. The agent must learn to optimize this policy in order to maximize the long-term expected discounted reward that it obtains. 

Policy gradient methods systematically adjust the policy parameters in order to maximize this objective. The policy parameters affect the sampled trajectory, which in turn affects the expected reward. A set of weights $\theta$ parameterizes the policy $\pi$. Thus, we can write the expected reward in terms of the policy parameters as follows:

\begin{equation}
    J(\theta) = \mathbb{E}_{p(S_{1:T};\theta)} \sum_{t=1}^{T} r_{t} = \mathbb{E}_{p(S_{1:T};\theta)} [R]
\end{equation}

The policy gradient theorem \cite{sutton1998introduction} gives us the gradient of the objective: 

\begin{equation}
    \nabla_{\theta} J(\theta) = \mathbb{E}_{p(S_{1:T};\theta)} \sum_{t=1}^{T} \nabla_{\theta} log[\pi(a_{t}|s_{1:t};\theta)] R_{t}
\end{equation}

In the above equation, $R_t$ is the discounted return obtained from state $s_t$ onwards. The REINFORCE algorithm \cite{sutton2000policy} uses the discounted return with a optionally subtracted baseline to reduce variance. In this work, we use the A3C framework \cite{MnihA3C}, which uses a critic to estimate the state-value function and updates networks for the policy and value functions in an asynchronous manner. 

\section{Related Works}
\label{sec:related}
Our work is related to a family of hierarchical RL methods \cite{barto2003recent, dietterich2000hierarchical, konidaris2007building, konidaris2012robot, konidaris2016constructing, kulkarni2016hierarchical}. Approaches in hierarchical RL typically learn the subgoal policies and a meta-policy simultaneously, using intrinsic rewards for completion of subgoals \cite{kulkarni2016hierarchical} or by tying parameters across different modules \cite{andreas2016modular} or by adopting a meta-learning approach \cite{frans2017meta}. A fundamental difference in our approach is that instead of learning optimal decompositions for a given complex task, we take the view of learning optimal compositions given a set of base tasks. The act of learning to grill a pancake \cite{frans2017meta-blog} does not require us learn an optimal, \textit{sequential} decomposition of cooking by interacting with a wide variety of recipes. Instead, it is much easier if a base set of skills, such as whisking eggs, measuring flour, heating the pan etc. can be composed to occur simultaneously, in sequence, optionally or held true until another subgoal is satisfied. The advantage of the ComposeNet architecture is that the overall tasks can be constructed post-hoc and pre-learned policies for skills can be quickly composed together to solve unseen tasks. This achieves much greater re-use of skills and quicker transfer to composed tasks as the skill networks are frozen after training once. Oh et al. \cite{oh2017zero} describe a framework to optimally sequence skills that can be learned in isolation from the main task. But they limit their discussion to sequence of subgoals, like program instructions, with occasional interruptibility for a higher priority task.

Our work is also similar to a related framework  in hierarchical RL called options \cite{SuttonOptions, bacon2017option}. The key difference here is that the skills are not provided to the agent as augmentations of its action set. Instead, our model learns skill-state embeddings, which are provided to a composition function which then learns to aggregate them and output an embedding for the overall task. Typically, multiple options cannot be activated in parallel. At a given state, an agent may activate a legal option and chose an action according to the policy prescribed by the option. After choosing the option, it must follow it for at least one step before activating another option or a taking a primitive action. In contrast, by composing skill-state embeddings, the agent is able to arbitrate between multiple sub-policies simultaneously to form optimal behavior according to the composed task.

In this sense, modular reinforcement learning is a closer analogue to our approach \cite{SAMEJIMA2003985, uchibe1996behavior, bhat2006difficulty, andreas2016modular}.  The skills can be regarded as sub-modules and the compositional layer as an aggregator that combines each skill's suggestion into a policy for the overall task. A crucial distinction in our work is that the skill modules do not provide direct policy recommendations to the aggregator and nor does the aggregator output a policy. Instead, they both learn to create skill-state embeddings for their particular skills or tasks. A final layer transforms embeddings in this space into policy actions. Representing submodule policies with embeddings allows us to create a richer description of the state, conditioned on each skill, in a way that allows us to create a trainable composition function.

% Also related are ensemble methods in RL \cite{romoff12separation, laroche2017multi, van2016improving}. Dietterich discusses ways to create ensembles using techniques such as bagging and error correcting output coding \cite{dietterich2000ensemble}. The composition function can be regarded as an aggregator for ensembles of skills as in \cite{wiering2008ensemble}, with the distinction that it is a function that can be trained using samples from the environment, its inputs are embeddings of skills rather than the policies themselves, and compositions can be arranged into hierarchies. Wiering and van Hasselt \cite{wiering2008ensemble} discuss ensemble aggregators such as majority voting, i.e. picking the action suggested the most, and Boltzmann multiplication, i.e. multiplying the ensembles' policies to form a distribution over actions. Although they suggest these in the context of ensemble algorithms, we use Boltzmann multiplication as a baseline in our experiments.  

Also related is work in multi-task RL, such as by van Seijen et al. \cite{van2017hybrid}, who use a Hybrid Reward Architecture agent to solve the game of Ms Pacman. Our work can be seen as lying between multi-task and hierarchical RL as our framework is capable of solving simultaneous goals, sequential goals and also optional goals, goals that must be held true until other goals are satisfied, etc. 

\section{ComposeNet}
\label{sec:architecture}
\begin{figure}[t]
  \centering
  \includegraphics[width=0.8\linewidth]{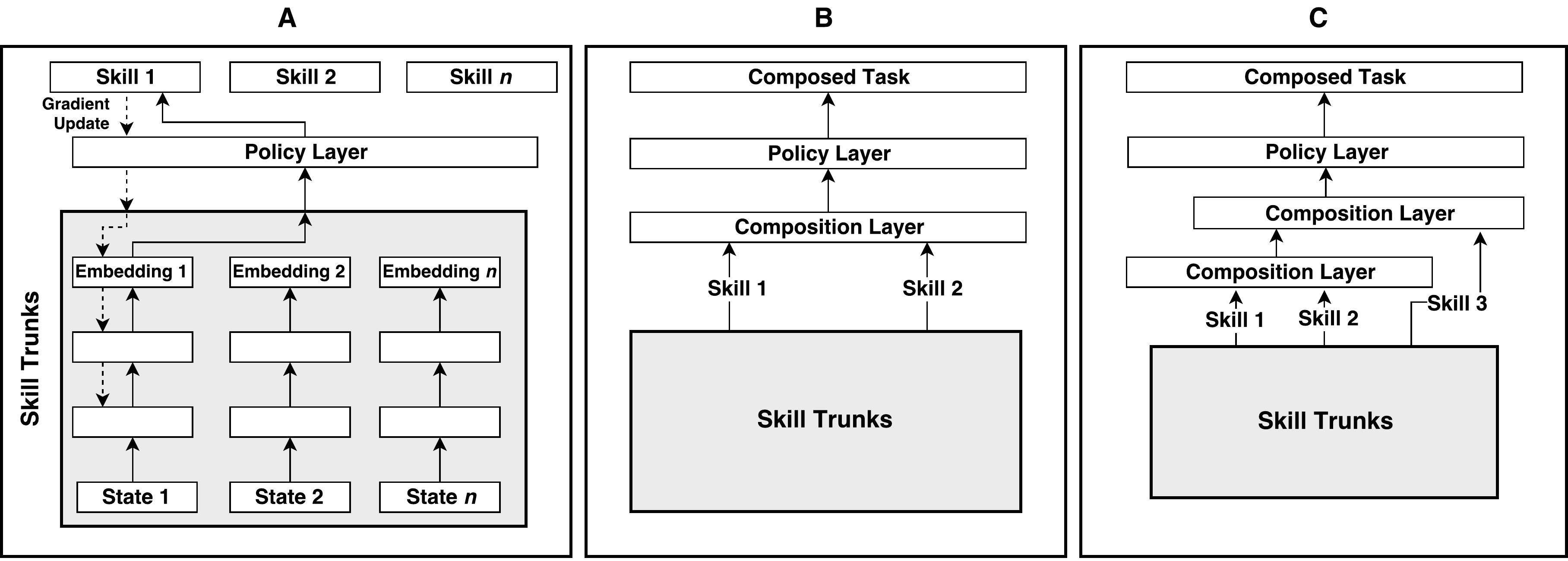}
  \caption{The ComposeNet architecture. (a) Training of the base skills. Each skill has its own trunk but shares a final layer, the policy layer, with all other skills. (b) Once the skill trunks and policy layer have been trained, skill-state embeddings are concatenated and fed into the composition layer. (c) By composing compositions of skills with other skills, one can create complex hierarchies.}
  \label{fig:architecture}
\end{figure}
The ComposeNet architecture allows learning of skill-state embeddings which can be used inside a differentiable composition function. Each skill has its own network trunk but the final layer, called the \textit{policy layer}, is shared across all the skills (figure \ref{fig:architecture}a). Each trunk is trained for its particular skill in isolation but gradients from all the skills are applied to the policy layer. The trunks are therefore forced to encode information about their particular skill as well as the agent state in their topmost layer. The policy layer is learning to take embeddings from any skill trunk and output a policy corresponding to that skill. This can be seen as a reversal of many multi-task learning architectures where a common input trunk is used with branches at the top for different tasks \cite{van2017hybrid}. In that case, a common embedding is learned for all tasks. Our goal is the opposite, i.e. to learn unique embeddings for each task and a common layer that can take any embedding and output the corresponding policy.

Now that we have a way to embed skill and state information in a single vector, we can combine two or more embeddings to create a new embedding for a composition of those skills. A composition, then, is a mapping from embeddings of all relevant skills to an embedding of the composed task.

\begin{equation}
C \colon \langle \iota_{e}^{(1)}(S), \iota_{e}^{(2)}(S), \ldots \iota_{e}^{(n)}(S) \rangle \to \iota_{c}(S),
\end{equation}

where $\iota_{e}^{(i)}(s)$ is an embedding for skill $i$ in state $s$ and $\iota_{c}(S)$ is an embedding for the composed task. Policies are formed by mapping embeddings to a probability distribution over actions.

\begin{equation}
\pi_{k} \colon \iota_{e}^{(k)}(S) \to p(A)
\end{equation}

Note that the policy function is agnostic to where the embeddings are coming from. This means that the same function must learn to map embeddings of all skills and any of their compositions to a policy. This property allows us to do recursive compositions of composed embeddings with other skills and create hierarchies of behavior. If both functions are differentiable, gradients with respect to the parameters of the composition function, $C$, can be formed using gradient based RL methods (see section \ref{background}).

In practice, two skill-state embeddings are concatenated end-to-end and fed to a fully connected layer, or the \textit{composition layer}, which acts as the composition function $C$. The output of the composition layer is the same dimensionality as the skill-state embeddings and is fed into the pre-trained policy layer whose output is now treated as a policy for the composed task (figure \ref{fig:architecture}b). Hence, the composition layer must learn to take two skill-state embeddings and output an embedding for the composed task. It is assumed that the correct skills for the task are provided to the agent and the form of the composition is known. This can be seen as a semi-supervised way of representing the task.

ComposeNet is trained as follows. First, skill trunks and the shared policy layer (without the composition layer) are trained simultaneously using asynchronous actor critic (A3C) \cite{MnihA3C}. Once converged, the weights of the skill trunks and the policy layer are frozen. Now a task consisting of a composition of two or more skills is chosen and only the composition layer is trained on samples from it.

\section{Environment and Skills}

To test our approach we devised a domain similar to Pacman, where an agent must collect or evade colored objects, Red, Green and Blue. The objects to collect remain stationary but the enemies chase the agent along the shortest path. Once an object is collected, it disappears from the map. The agent can teleport across the map if it goes out of bounds, but the objects cannot. An example task in this environment is ``collect object Red while evading object Blue'' ($\neg b \; \mathcal{U} \; r$). The agent's state is a 15x15 pixel grayscale image of the game grid. There are six skills in this environment: collecting and evading the three objects respectively. The skills are trained separately in environments with reward functions only relevant to that skill.

We consider four types of compositions in this environment,
\begin{enumerate}
\item $\neg p \; \mathcal{U} \; q$, collect object $q$ while evading enemy $p$; 
\item $\lozenge p \, \vee \, \lozenge q$, collect object $p$ or $q$;
\item $\square \, \neg p \, \wedge \, \square \, \neg q $, always evade enemy $p$ and enemy $q$; and
\item $\lozenge ( p \, \wedge \, \lozenge q )$, collect object q then object p.
\end{enumerate}

It should be noted that the environment does not provide any explicit reward signal indicating the relevance of a subgoal to the current state or when it is complete. A reward is only issued when the full task is complete.

\section{Results}

We first train all six skills networks for about 3 million steps total (i.e. 500,000 steps on average per skill). After this, skill networks and the policy layer are frozen. This initial cost is fixed and amortized over all possible compositions.

In the following graphs, we compare performance of our method (ComposeNet) to two baselines: (1) training a single network from scratch, and (2) a meta-controller approach where a network picks from relevant trained skills every step. We also experimented with using skills as options by augmenting the agent's action space. That performed worse than training from scratch on all problems, likely due to the increased number of actions. We have omitted those results for clarity.   

\subsection{Single Compositions and Zero-Shot Generalizations}
\begin{figure}[t]
\centering
\begin{subfigure}{.40\textwidth}
  \centering
  \includegraphics[width=\linewidth]{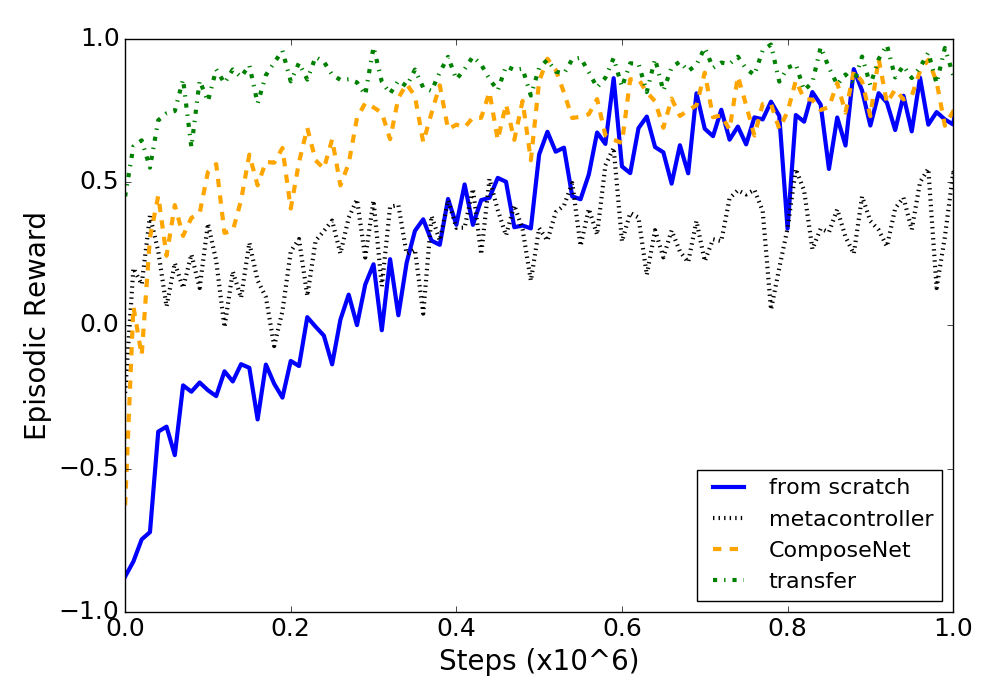}
  \caption{$\neg g \, \mathcal{U} \, b$}
  \label{fig:collect_2_evade_1}
\end{subfigure}
\begin{subfigure}{.40\textwidth}
  \centering
  \includegraphics[width=\linewidth]{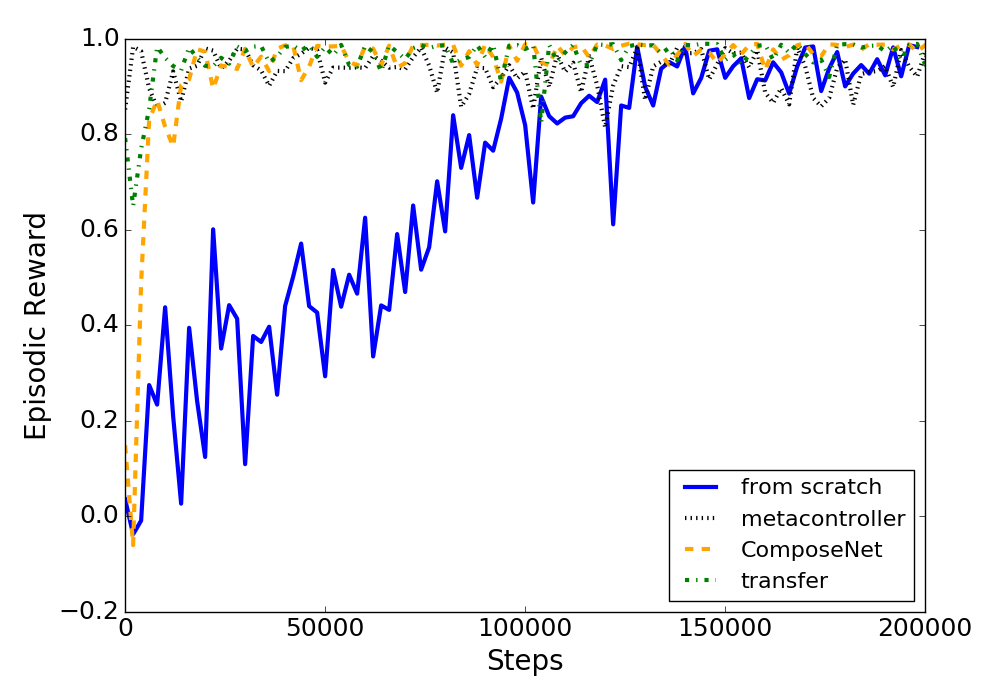}
  \caption{$\lozenge r \, \vee \, \lozenge b$}
  \label{fig:collect_0_or_collect_2}
\end{subfigure}

\centering
\begin{subfigure}{.40\textwidth}
  \centering
  \includegraphics[width=\linewidth]{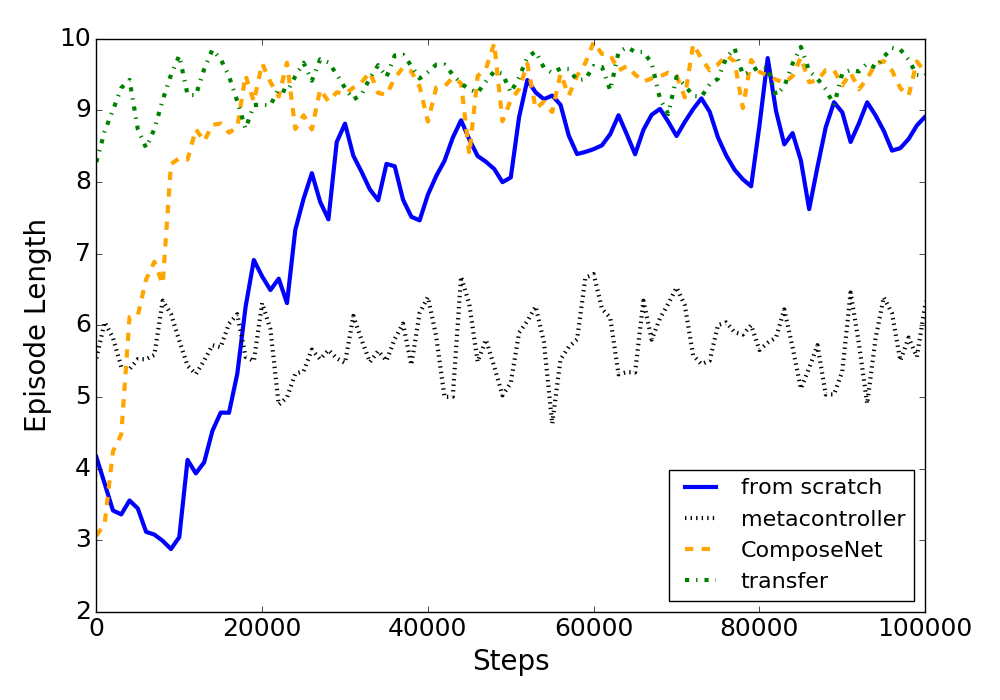}
  \caption{$\square \, \neg r \, \wedge \, \square \, \neg g $}
  \label{fig:evade_0_and_evade_1}
\end{subfigure}
\begin{subfigure}{.40\textwidth}
  \centering
  \includegraphics[width=\linewidth]{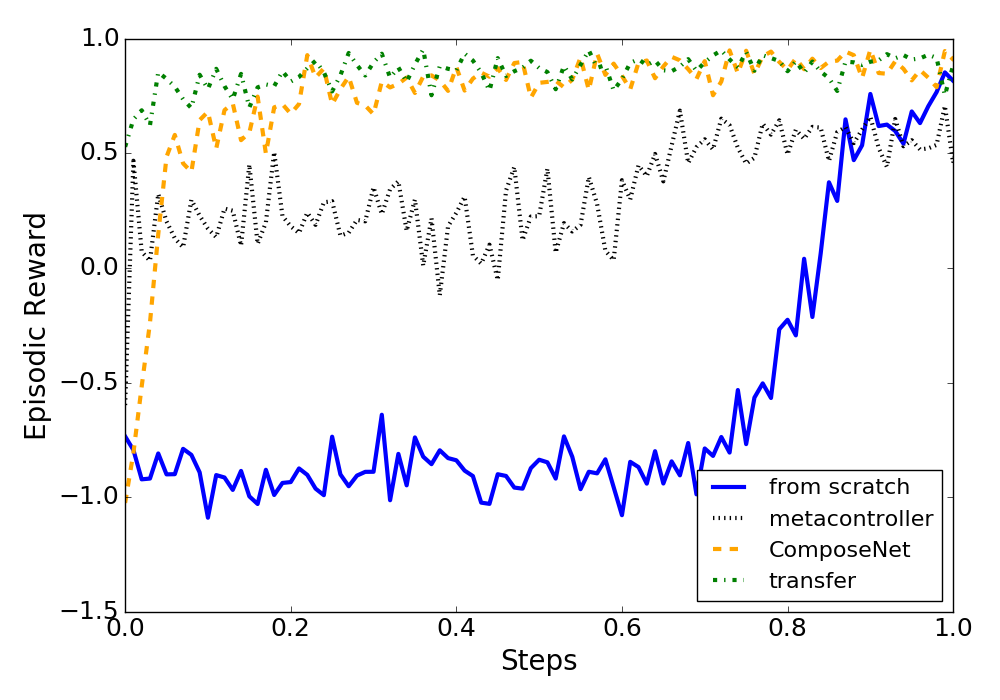}
  \caption{$\lozenge ( r \, \wedge \, \lozenge g )$}
  \label{fig:collect_0_then_collect_1}
\end{subfigure}
\caption{Performance of our method (ComposeNet) compared with baselines of metacontroller strategy and learning from scratch. In the `transfer' condition, the composition layer is initialized with weights learned over similar tasks. (a) \textbf{Collect blue while evade green} zero-shot reward: 0.45. (b) \textbf{Collect red or blue} zero-shot reward: 0.79. (c) \textbf{Evade red and green} zero-shot episode length: 8.28 (d) \textbf{Collect red then green} zero-shot reward: 0.53.}
\label{fig:compose_objects}
\end{figure}
Figure \ref{fig:compose_objects} shows performance of ComposeNet compared to our baselines on a sample task for each type of composition. Overall, the results show that individual skills can be successfully composed with the ComposeNet architecture to near optimality, and are learned faster than either of the baselines. For the ``while'' and ``then'' compositions (figures \ref{fig:collect_2_evade_1} and \ref{fig:collect_0_then_collect_1}), the meta-controller initially achieves a somewhat good reward but then learning slows down significantly. This is because the meta-controller quickly learns that the skill ``collect blue'' may lead to a high reward. But improving the reward requires it learn to alternate between reaching blue and evading green. Similarly with ``collect red then green'', the agent may reach red then randomly stumble into green, or follow the ``collect green'' skill only and collect red along the way accidentally. But activating them in sequence with correct timing is harder to learn. The exception is ``collect red or blue'', as in this case the meta-controller can select any skill at random and ensure a high reward. ComposeNet quickly learns to achieve high reward for all types of compositions. For the $p \, \wedge \, q$ composition, the meta-controller strategy completely fails to learn. Evading both objects is a hard task and actions must be chosen to evade both at the same time. Activating only one, say `avoid red', may lead the agent towards danger, towards green. Our learned composition function ensures both are evaded simultaneously. This is an example of why the ComposeNet architecture is better suited to a wide variety of compositions than traditional hierarchical RL approaches.

We also tested zero-shot task generalization by training the composition layer on other tasks containing the same compositions. For example, we trained the same composition layer on all five tasks of the type $\neg p \, \mathcal{U} \, q$, except $\neg g \, \mathcal{U} \, b$. The learned weights were then used as initialization for the composition layer of the held out task. Our results show that there is significant zero-shot generalization to compositions of the same type. Further training on the held out task quickly produces near optimal rewards. For the ``collect red or blue'' and ``evade red and green'' tasks we transferred from only two other tasks, as the order of the objects does not matter in these compositions.

\subsection{Hierarchies of Compositions}
\begin{figure}[t]
\centering
\begin{subfigure}{.4\textwidth}
  \centering
  \includegraphics[width=\linewidth]{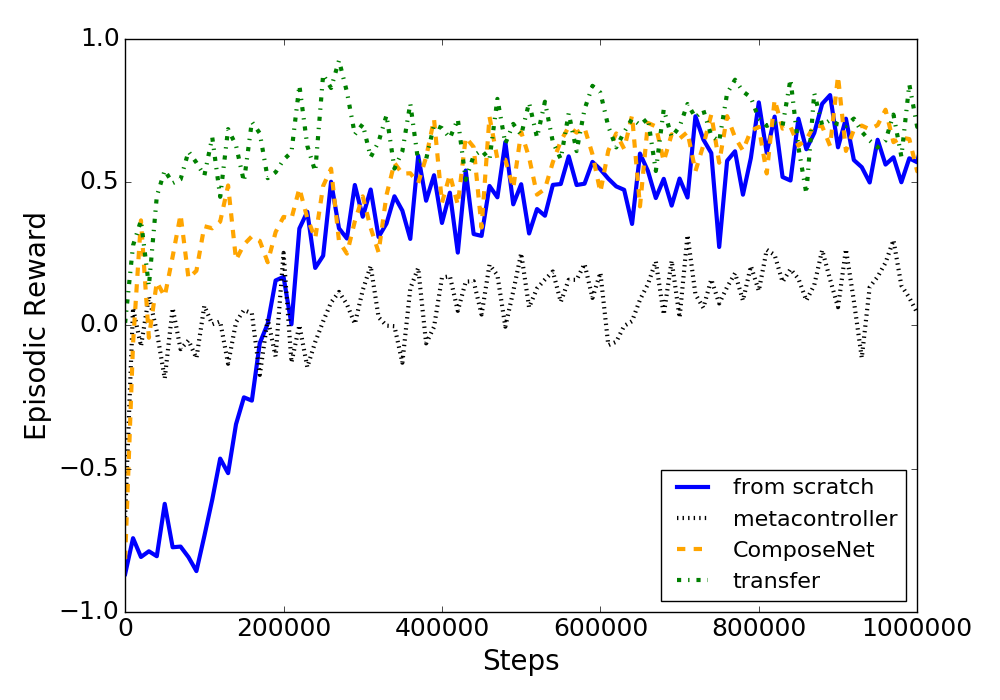}
  \caption{$(\neg g \, \wedge \, \neg b) \, \mathcal{U} \, r$}
  \label{fig:target_0_avoid_1_avoid_2}
\end{subfigure}
\begin{subfigure}{.4\textwidth}
  \centering
  \includegraphics[width=\linewidth]{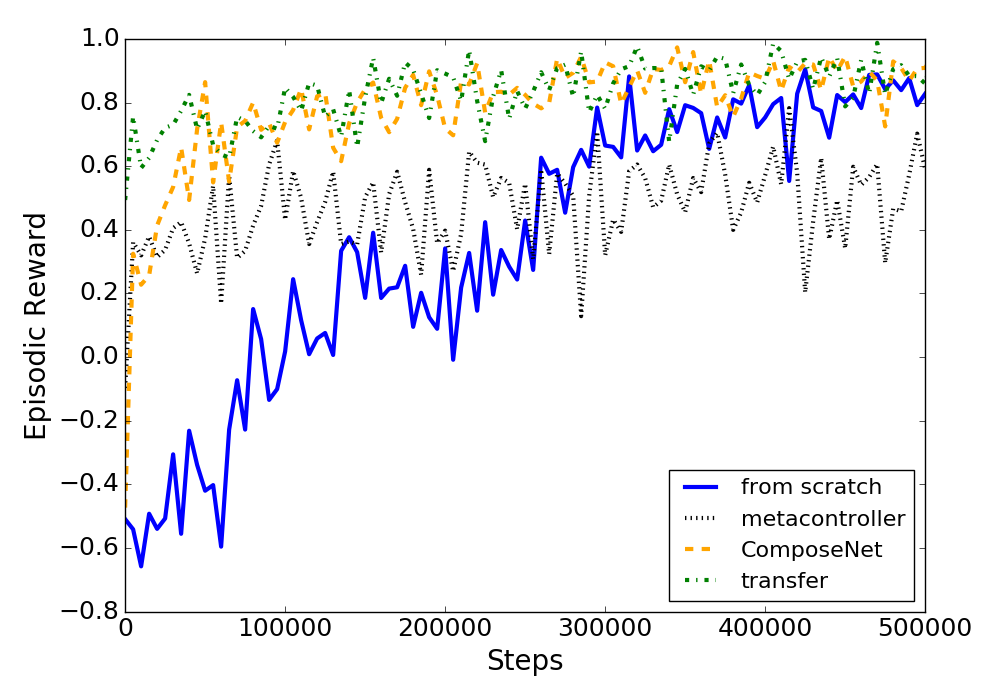}
  \caption{$\neg b \, \mathcal{U} \, (r \, \vee \, g)$}
  \label{fig:target_0_1_avoid_2}
\end{subfigure}
\caption{Hierarchies of compositions allow for even more complex tasks to be learned quickly. (a) Zero-shot reward: 0.01. The task of evading two enemies while collecting the third object is challenging and success rate is low for all policies. ComposeNet reaches a good rate of success fairly quickly. (b) Zero-shot reward: 0.49. In this case, the task was of collecting either one of two objects while evading the third.}
\label{fig:hierarchy}
\end{figure}

The ComposeNet architecture is versatile enough that the composition layer can accept itself as an input, leading to more complex hierarchies. Figure \ref{fig:hierarchy} shows results on two composed tasks, `collect red while evade green and blue', $(\neg g \, \wedge \, \neg b) \, \mathcal{U} \, r$, and `collect red or green while evade blue', $\neg b \, \mathcal{U} \, (r \, \vee \, g)$. The networks are formed by first composing the literals in parentheses, and then composing the resulting embedding with the embedding for the third literal. For example, in figure \ref{fig:target_0_avoid_1_avoid_2}, the embeddings for `evade green' and `evade blue' are composed first. The output is fed into another composition layer, along with the embedding for `collect red'. The output of this layer is then fed to the policy layer. In the `transfer' condition for this task, the first composition layer was initialized with weights trained on all `evade this and that' tasks. These weights have been trained to compose two evade policies into a single policy that successfully evades both objects. The second composition layer is initialized with weights trained on all `collect this while evade that' tasks. This layer takes as input `collect red' embeddings and the composed embedding from the first compositional layer and produces an embedding for the complete task. Similarly, for the transfer treatment in the second task, weights from training on all `collect this or that' tasks and all `collect this while evade that' tasks were used to initialize the two composition layers. 

The results for zero-shot generalization show that some transfer occurs to such hierarchically composed tasks, even when the training set is comprised solely of flat compositions of two literals. The task in figure \ref{fig:target_0_avoid_1_avoid_2} is fairly challenging, so the zero-shot policy is able to collect reward only about half the time, resulting in an average reward close to zero. With a few samples from the composed task, it quickly learns a high-reward policy. In figure \ref{fig:target_0_1_avoid_2}, the transferred policy starts with a decent zero-shot reward of close to 0.5 and also converges quickly. ComposeNet allows for this mix-and-match composition capability and reuse of learned skills, even in complicated hierarchies.

\subsection{Ablative Studies}
\label{sec:analysis}
\begin{figure}[h]
\centering
\begin{subfigure}{.45\textwidth}
  \centering
  \includegraphics[width=\linewidth]{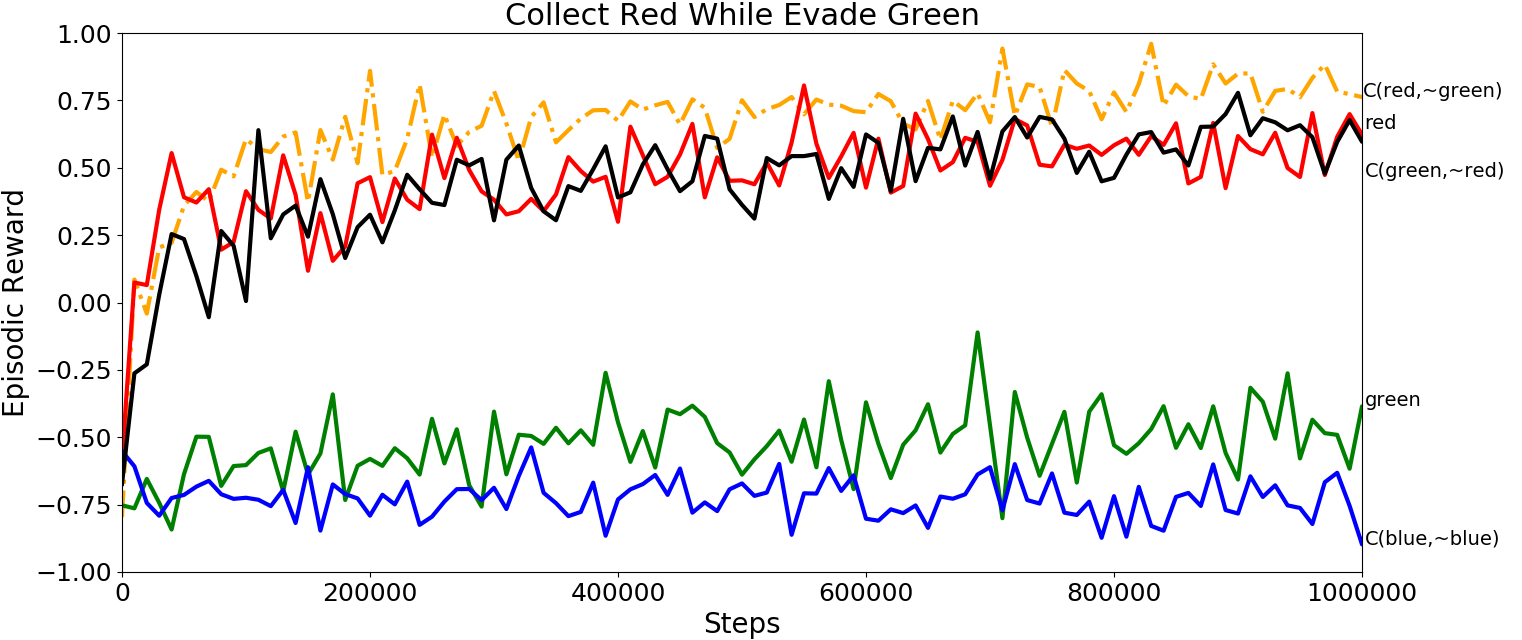}
%   \caption{}
\label{fig:target_0_avoid_1_wrong_skills}
\end{subfigure}
\begin{subfigure}{.45\textwidth}
  \centering
  \includegraphics[width=\linewidth]{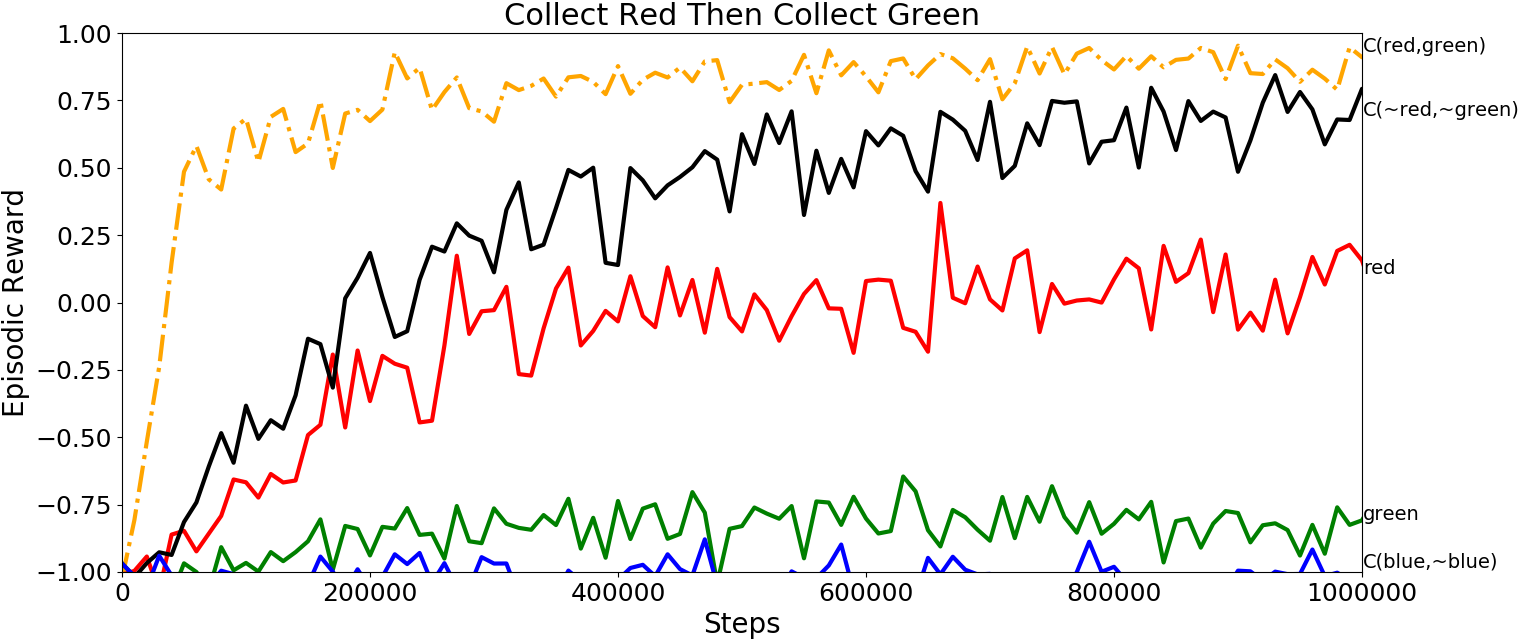}
%   \caption{}
  \label{fig:target_0_then_1_wrong_skills}
\end{subfigure}
\caption{Degradation of performance by losing the compositional layer or by using the wrong skills. $C$ denotes the composition function and the labels with only skills denotes the case where a fresh policy layer is trained on top of that skill trunk. Tasks are learned best with the correct policies composed and do not learn at all when no relevant information on objects of interest can be gleaned from the skill-embedding.}
\label{fig:ablative}
\end{figure}

We tested two ablative conditions to determine the importance of two main components of the ComposeNet architecture. These conditions offer additional support for the argument that skill-state embeddings are able to represent useful information pertaining to the skill and are important for creating a trainable composition function.

\textbf{Incorrect skills}. In this condition, skill-embeddings from the wrong skills are passed to the composition function and training occurs as normal. These are denoted by the $C(p,q)$ labels in figure \ref{fig:ablative}, where $p$ and $q$ are individual skills. Overall, the task is learned the best when the correct skills are composed and performance drops significantly when incorrect skills are used and almost no learning occurs from skills of irrelevant objects.

For the ``collect red while evade green'' task, composing the `collect red' and 'evade green' skills gives the best performance. But composing `collect green' and `evade red' skills learns a decent policy as well. This occurs because the `evade red' skill provides information on the red object, specifically on how to evade it. By reversing the policy for evading the red object, one can collect it instead. The same is almost true for reversing the `collect green' skill as well, although the opposite of collecting an object may not be the optimal policy for evading it in all cases. When composing two skills that provide no information on either of the pertinent objects, `collect blue' and `evade blue', no learning occurs. The same behavior can be seen with the other task; composing the correct skills leads to the highest reward; composing skills that provide some information about the objects of interest leads to lower, but still positive, rewards; and composing two skills with no pertinence to the objects of interest does not learn at all. This shows that skill-state embeddings do encode relevant information for the task and any arbitrary representation will not work just as well.

\textbf{No compositional layer}. We also tested whether the composition function truly requires embeddings from both the skills. Can an equally good policy be learned by training a fresh policy layer on top of a pre-trained skill trunk using samples from the composed task? In these experiments, only the policy layer's weights are updated. Weights of the skill trunks are held fixed. Results are shown in figure \ref{fig:ablative}, with labels as the names of the skills trunks on which the policy layer was re-trained. For example, in the ``collect red while evade green'' task, a fresh policy layer was learned on top of the trained skill trunk for the `collect red' skill. This performs slightly worse than composing the right skills, likely because it gets caught by the green object more frequently. For the ``collect red then green'' task, the same architecture performs much worse than using ComposeNet. It likely is only able to guide the agent to the red object, which nets no reward in the composed task. Training a fresh policy layer on top of the `collect green' skill performs even worse for this task, because if the agent collects the green object first, the task is impossible to complete correctly. Once again, we have shown that this component of the ComposeNet architecture is crucial for learning the task effectively as a blending of both skills is necessary to learn all types of composed tasks.

\section{Conclusion and Future Work}
We have presented a framework called ComposeNet which allows an agent to compose simple skills into a hierarchy to solve complicated tasks. The skills are learned separately and can be reused for multiple compositions. Key in the framework are skill-state embeddings and a trainable composition function, backed by our ablative studies. Moreover, when testing on composed tasks it has never seen before, ComposeNet shows some zero-shot generalization capability, and quickly converges with few environment samples. This suggests that the ability to compose skills in this domain may be transferable. Future work in this area includes trying ComposeNet on more complicated domains such as Minecraft. We have demonstrated that the operators: (1) $\mathcal{U}$ (\textit{collect this while evade that}), (2) $\wedge$ (\textit{evade this and that}), (3) $\lor$ (\textit{collect this or that}), and (4) $\lozenge(\, \wedge \, \lozenge)$ (\textit{collect this then that}), can be learned quickly with pre-trained base skills. Work on learning other types of compositions is ongoing.
\bibliography{nips_2017}
\bibliographystyle{abbrv}
\end{document}